\documentclass{article}
\usepackage{spconf,amsmath,graphicx,hyperref}
\usepackage{bm}
\usepackage{amsfonts}
\usepackage{multirow}
\usepackage[table]{xcolor}
\usepackage{pifont}
\newcommand{\cmark}{\textbf{\textcolor{green}{\ding{51}}}} 
\newcommand{\xmark}{\textbf{\textcolor{red}{\ding{55}}}}   

\title{WaveletGaussian: Wavelet-domain Diffusion for Sparse-view 3D Gaussian Object Reconstruction}
\name{Hung Nguyen, Runfa Li, An Le, Truong Nguyen}
\address{Video Processing Lab, UC San Diego}
\begin{document}
%
\maketitle

\begin{figure*}[t]
    \centering \includegraphics[width=\linewidth]{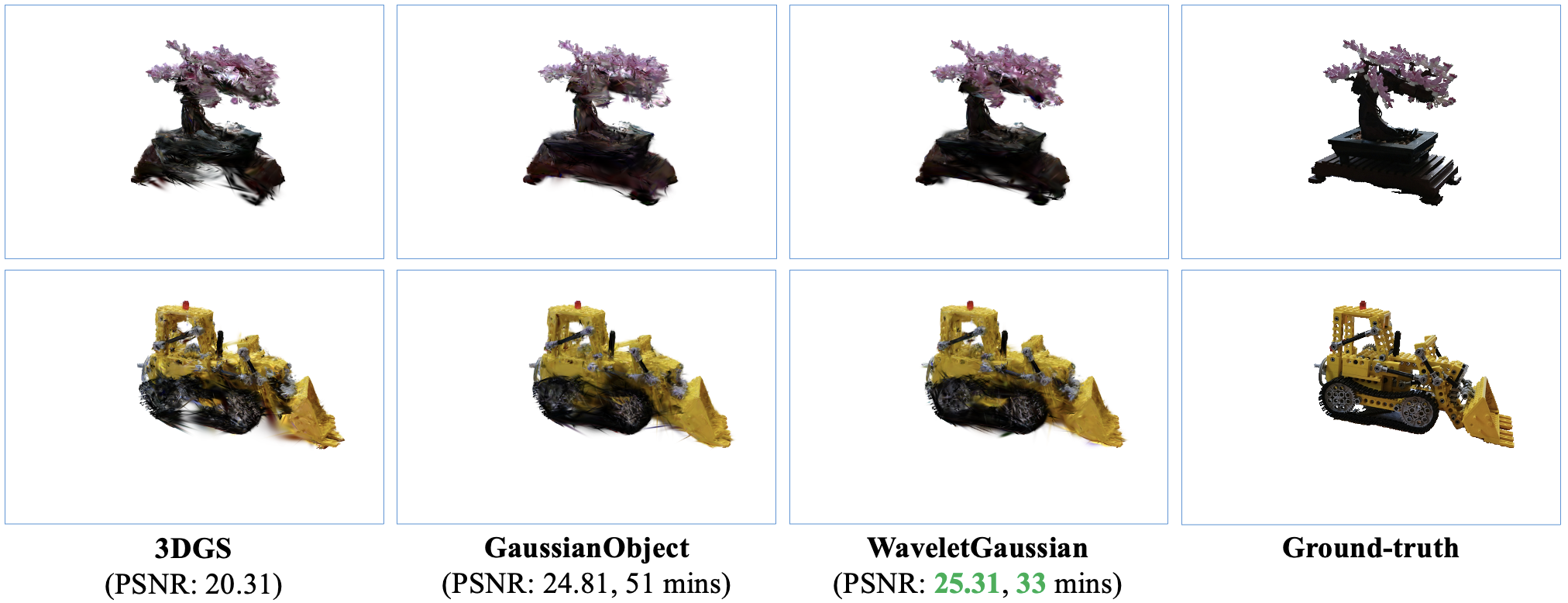}
    \caption{We propose WaveletGaussian, a framework for sparse-view 3D Gaussian object reconstruction based on wavelet-domain diffusion model repair, which significantly reduces training time while bettering rendering quality.}
    \label{fig_quant}
\end{figure*}

\begin{abstract}
3D Gaussian Splatting (3DGS) has become a powerful representation for image-based object reconstruction, yet its performance drops sharply in sparse-view settings. Prior works address this limitation by employing diffusion models to repair corrupted renders, subsequently using them as pseudo references for later optimization. While effective, such approaches incur heavy computation from the diffusion fine-tuning and repair steps. We present WaveletGaussian, a framework for more efficient sparse-view 3D Gaussian object reconstruction. Our key idea is to shift diffusion into the wavelet domain: diffusion is applied only to the low-resolution LL subband, while high-frequency subbands are refined with a lightweight network. We further propose an efficient online random masking strategy to curate training pairs for diffusion fine-tuning, replacing the commonly used, but inefficient, leave-one-out strategy. Experiments across two benchmark datasets, Mip-NeRF 360 and OmniObject3D, show WaveletGaussian achieves competitive rendering quality while substantially reducing training time.
\end{abstract}

\section{Introduction}
\label{sec:intro}

3D Gaussian Splatting (3DGS) \cite{3DGS} has become a leading approach for reconstructing 3D scenes or objects from 2D images, producing photorealistic novel views with relatively short training times. Nevertheless, it generally depends on densely captured training views with accurate camera poses, which demand significant effort in data collection. In scenarios with sparse views, the reconstructed geometry is poorly constrained, often leading to artifacts or unstable structures that severely degrade rendering quality. This limitation reduces its practicality in real-world settings, where acquiring dense, well-posed data is often impractical \cite{GaussianObject}.

Therefore, sparse-view 3DGS has emerged as an active research direction. While multiple kinds of priors have been leveraged for the task \cite{DWTGS}, denoising diffusion models (DDMs) \cite{DDM} have emerged as a powerful option due to their outstanding generative capabilities. Within a sparse-view 3DGS framework, they are often used to repair the renders from novel viewpoints, which are often highly corrupted due to the lack of explicit supervision. The repaired views are subsequently used as pseudo references for later optimization, thus emulating artifact-free dense-view training \cite{DeceptiveNeRF, 3DGS-Enhancer, Difix3d+, GaussianObject, Free360, GenFusion, RI3D, Generative3DGS}. Despite producing high-quality pseudo ground-truths, this approach incurs significant computation due to the required fine-tuning step, which is necessary to adapt a pre-trained diffusion model to the specific scene or object at hand. The repair step is also costly, thus severely hindering the method's scalability. To shorten the overall training time, recent works leverage LoRA \cite{LoRA} adapters, but a single scene can still take up to an hour to train \cite{GaussianObject}.

In this paper, we introduce the WaveletGaussian framework for 3D Gaussian object reconstruction under sparse views, aiming to significantly reduce overall training time while maintaining competitive rendering quality. To achieve this objective, WaveletGaussian proposes repositioning the diffusion fine-tuning and repair steps from the RGB to wavelet domain. The rationale is that the latter is only half-resolution, while still preserving all information through the lossless wavelet transform. Specifically, the diffusion model is only trained on, and applied to, the low-frequency LL subband, while the high-frequency subbands are processed using a lightweight U-Net-like \cite{U-Net} architecture. Additionally, we propose a novel online random masking method to curate the object-specific dataset for diffusion fine-tuning, replacing the commonly used, but inefficient, leave-one-out strategy \cite{RI3D, 4DSloMo, GaussianObject}. In summary, our contributions are as follows:

\begin{itemize}
    \item We propose WaveletGaussian, a 3DGS-based framework for sparse-view object reconstruction with significantly reduced training times due to i) wavelet-domain, diffusion-based novel view repairs, and ii) an efficient method to curate the object-specific dataset for diffusion fine-tuning.
    \item Through experiments on benchmark datasets, our WaveletGaussian demonstrates to significantly reduce overall training time, while maintain competitive rendering quality.
\end{itemize}

\section{Related Works}
\label{sec:related}

\textbf{Discrete Wavelet Transform (DWT) for 3DGS}. Recently, the DWT has attracted growing attention within deep computer vision frameworks, as it disentangles frequency learning while providing efficiency benefits \cite{DWTGS}. Extensions to 3DGS are also being explored, e.g., for fine detail enhancement \cite{3D-GSW}, coarse-to-fine efficient learning \cite{autoopti3dgs} and frequency regularization \cite{DWTGS}. Our WaveletGaussian novelly introduces the DWT to a sparse-view framework with diffusion-based repairs to improve efficiency.

\textbf{Diffusion-based repair for sparse-view 3DGS}. Denoising diffusion models (DDMs) \cite{DDM}, known for their strong generative capabilities, are widely used to repair the highly corrupted novel views of sparse-view 3DGS. However, this approach incurs significant computation, as it requires fine-tuning the diffusion model on large-scale datasets \cite{DeceptiveNeRF, 3DGS-Enhancer, Difix3d+, GenFusion}. Scene-specific fine-tuning and LoRA adapters \cite{GaussianObject, RI3D, Generative3DGS, Free360} improve efficiency, but the total training time may still require up to an hour \cite{GaussianObject}, thus severely limiting the method's scalability. Our WaveletGaussian proposes repositioning the diffusion-related processes to the lower-resolution wavelet domain for efficiency benefits.

\begin{figure*}[t]
    \centering \includegraphics[width=\linewidth]{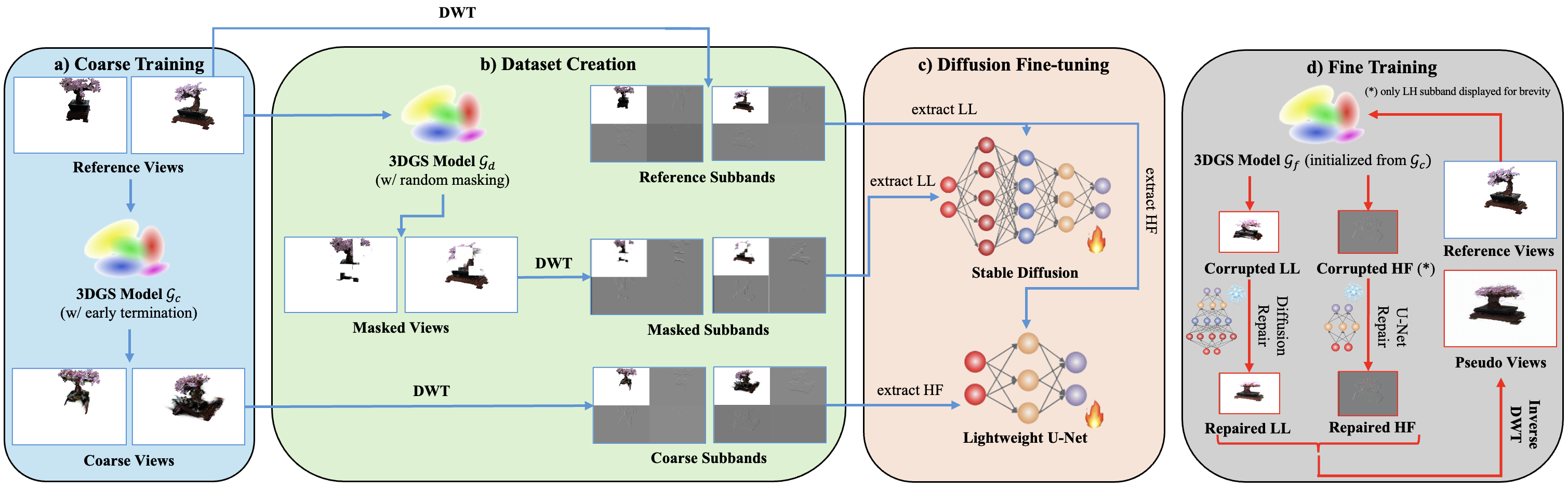}
    \caption{The proposed WaveletGaussian framework for sparse-view 3D Gaussian object reconstruction. Central to WaveletGaussian is repositioning of the diffusion model \cite{ControlNet} from the RGB to lower-resolution wavelet domain for novel view repairs.}
    \label{fig_arch}
\end{figure*}


\section{Methodology}
\label{sec:method}

\subsection{Discrete Wavelet Transforms} \label{sec:dwt}

Given a 2D image $\mathbf{X}$, the Forward DWT decomposes it into four distinct subbands (LL, LH, HL, HH) as follows:
\begin{equation} \label{eq_fdwt}
\begin{split}
\mathbf{X}_{\text{LL}} = \mathbf{L}_0\mathbf{X}\mathbf{L}_1, \quad
\mathbf{X}_{\text{LH}} = \mathbf{H}_0\mathbf{X}\mathbf{L}_1, \\
\mathbf{X}_{\text{HL}} = \mathbf{L}_0\mathbf{X}\mathbf{H}_1, \quad
\mathbf{X}_{\text{HH}} = \mathbf{H}_0\mathbf{X}\mathbf{H}_1
\end{split}
\end{equation}
where $\mathbf{L}_{(\cdot)}$ and $\mathbf{H}_{(\cdot)}$ are the low-pass and high-pass filtering matrices applied to either the columns or rows of $\mathbf{X}$, as indicated by the subscript $\{0, 1\}$. As an example, the low-pass, vertically filtering matrix $\mathbf{L}_0$, based on Haar wavelet \cite{wavelet-book}, is:
\begin{equation*}
\mathbf{L}_0 =
\begin{bmatrix}
\frac{1}{\sqrt{2}} & \frac{1}{\sqrt{2}} & 0 & 0 & 0 & \cdots \\
0 & 0 & \frac{1}{\sqrt{2}} & \frac{1}{\sqrt{2}} & 0 & \cdots \\
\vdots & \vdots & \vdots & \vdots & \vdots & \ddots
\end{bmatrix}
\end{equation*}
which is constructed by shifting the low-pass, averaging filter $[1/\sqrt{2}, 1/\sqrt{2}]$ along rows. The shifts imply downsampling (in this case, by 2). The high-pass matrix $\mathbf{H}_0$ is constructed similarly, using the high-pass, differencing filter $[-1/\sqrt{2}, 1/\sqrt{2}]$ instead. In Equation \eqref{eq_fdwt}, the LL subband results from low-pass filtering in both directions, retaining the coarse structure of the image. The LH and HL subbands result from applying a high-pass filter in one direction and a low-pass filter in the other, capturing horizontal and vertical information, respectively. The HH subband, high-pass filtered in both directions, emphasizes fine diagonal textures.

Given the four subbands, the Inverse DWT provides the reconstruction $\mathbf{\hat{X}}$ as follows:
\begin{equation} \label{eq_idwt}
\hat{\mathbf{X}} =
\tilde{\mathbf{L}}_0^\top \mathbf{X}_{\text{LL}} \tilde{\mathbf{L}}_1^\top +
\tilde{\mathbf{H}}_0^\top \mathbf{X}_{\text{LH}} \tilde{\mathbf{L}}_1^\top +
\tilde{\mathbf{L}}_0^\top \mathbf{X}_{\text{HL}} \tilde{\mathbf{H}}_1^\top +
\tilde{\mathbf{H}}_0^\top \mathbf{X}_{\text{HH}} \tilde{\mathbf{H}}_1^\top
\end{equation}
where the matrices used in the Forward and Inverse DWT are termed ``analysis'' and ``synthesis'', respectively. The Haar synthesis matrices, $\tilde{\mathbf{L}}_0$ and $\tilde{\mathbf{H}}_0$, are constructed using the synthesis filters $[1/\sqrt{2}, 1/\sqrt{2}]$ (low-pass) and $[1/\sqrt{2}, -1/\sqrt{2}]$ (high-pass). The ``Perfect Reconstruction'' condition, which occurs when $\mathbf{X} = \mathbf{\hat{X}}$ and implies no loss of information, is satisfied when specific relationships exist between the analysis–synthesis filter pairs \cite{wavelet-book}. 

\subsection{Overall Framework}

Figure \ref{fig_arch} shows an overview of our proposed WaveletGaussian. Firstly, in the \textit{Coarse Training} (a) stage, a 3DGS model $\mathcal{G}_{c}$ is trained on all $N$ sparse views for some limited iterations to capture the overall geometry. As the training of $\mathcal{G}_{c}$ is terminated early, the resulting renders, even from known viewpoints, are moderately corrupted. We pass both the reference and coarse renders into the Forward DWT for later uses.


\begin{table*}[t]
\centering
\caption{Quantitative results, 4-view Mip-NeRF 360 \cite{Mipnerf360} and OmniObject3D \cite{OmniObject3D} datasets}
\label{tab_quant}
\resizebox{\linewidth}{!}{%
\begin{tabular}{|c|c|c|c|c|c|c|c|c|}
\hline
\multirow{2}{*}{Method} 
& \multicolumn{4}{c|}{Mip-NeRF 360 \cite{Mipnerf360}} 
& \multicolumn{4}{c|}{OmniObject3D \cite{OmniObject3D}} \\ \cline{2-9}
 & PSNR ($\uparrow$) & SSIM ($\uparrow$) & LPIPS ($\downarrow$) & Time (mins) 
 & PSNR ($\uparrow$) & SSIM ($\uparrow$) & LPIPS ($\downarrow$) & Time (mins) \\ \hline
3DGS \cite{3DGS} & 20.31 & 0.899 & 0.108 & -- & 17.29 & 0.930 & 0.086 & -- \\ 
FSGS \cite{FSGS} & 21.07 & 0.910 & 0.095 & -- & 24.71 & 0.955 & 0.063 & -- \\ 
GaussianObject \cite{GaussianObject} & 24.81 & 0.935 & 0.050 & 51 & 30.89 & 0.976 & 0.030 & 55 \\ 
WaveletGaussian (Ours) & \textbf{25.31} & \textbf{0.939} & \textbf{0.047} & \textbf{33} & \textbf{31.22} & \textbf{0.983} & \textbf{0.028} & \textbf{35} \\ 
\hline
\end{tabular}}
\end{table*}

\begin{table}[t]
\centering
\caption{Ablation studies on the 4-view Mip-NeRF 360 \cite{Mipnerf360} dataset with (\cmark) or without (\xmark) proposed components.}
\label{tab_abs}
\resizebox{\columnwidth}{!}{%
\begin{tabular}{|c|c|c|c|c|c|c|c|}
\hline
Offline RM & Online RM & wavelet-$\mathcal{D}$ & $\mathcal{U}$ repair & PSNR ($\uparrow$) & SSIM ($\uparrow$) & LPIPS ($\downarrow$) & Time (mins) \\ \hline
\xmark & \xmark & \xmark & \xmark & 24.81 & 0.935 & 0.050 & 51 \\
\cmark & \xmark & \xmark & \xmark & 24.95 & 0.934 & 0.051 & 43 \\ 
\xmark & \cmark & \xmark & \xmark & 25.10 & 0.934 & 0.051 & 43 \\ 
\xmark & \cmark & \cmark & \xmark & 24.99 & 0.934 & 0.051 & \textbf{30} \\
\xmark & \cmark & \cmark & \cmark & \textbf{25.31} & \textbf{0.939} & \textbf{0.047} & 33 \\ \hline
\end{tabular}}
\end{table}

The \textit{Dataset Creation} (b) stage involves synthesizing corrupted–clean image pairs to fine-tune a pre-trained diffusion model \cite{StableDiffusion} $\mathcal{D}$ and a lightweight U-Net-like \cite{U-Net} model $\mathcal{U}$. The fine-tuning is necessary to adapt them to object-specific details, enabling later repairs of novel views. To simulate corrupted patterns for $\mathcal{D}$, a 3DGS model $\mathcal{G}_{d}$ is optimized with a masking strategy, to be detailed in Section \ref{sec_orm}. The masked renders are paired with the reference ones, both transformed into the wavelet domain, where we retain only the LL subbands to form the LL-domain diffusion dataset. On the other hand, the corrupted patterns for $\mathcal{U}$ are retrieved from the high-frequency (HF) subbands (LH, HL and HH) of the coarse renders of $\mathcal{G}_c$, also paired with the corresponding references.

The \textit{Diffusion Fine-Tuning} (c) stage operates in the low-resolution LL domain. Here, $\mathcal{D}$ is essentially trained to be an inpainting model operating in low frequencies (LF), while $\mathcal{U}$ repairs the HF. By training separate models for LF/HF repairs, we disentangle frequency learning, allowing each model to specialize in LF/HF. Since both models operate at half resolutions, this remains considerably cheaper than fine-tuning a single RGB-domain $\mathcal{D}$, as will be shown in Section \ref{sec_abs}.

\begin{figure}[t]
    \centering \includegraphics[width=0.9\linewidth]{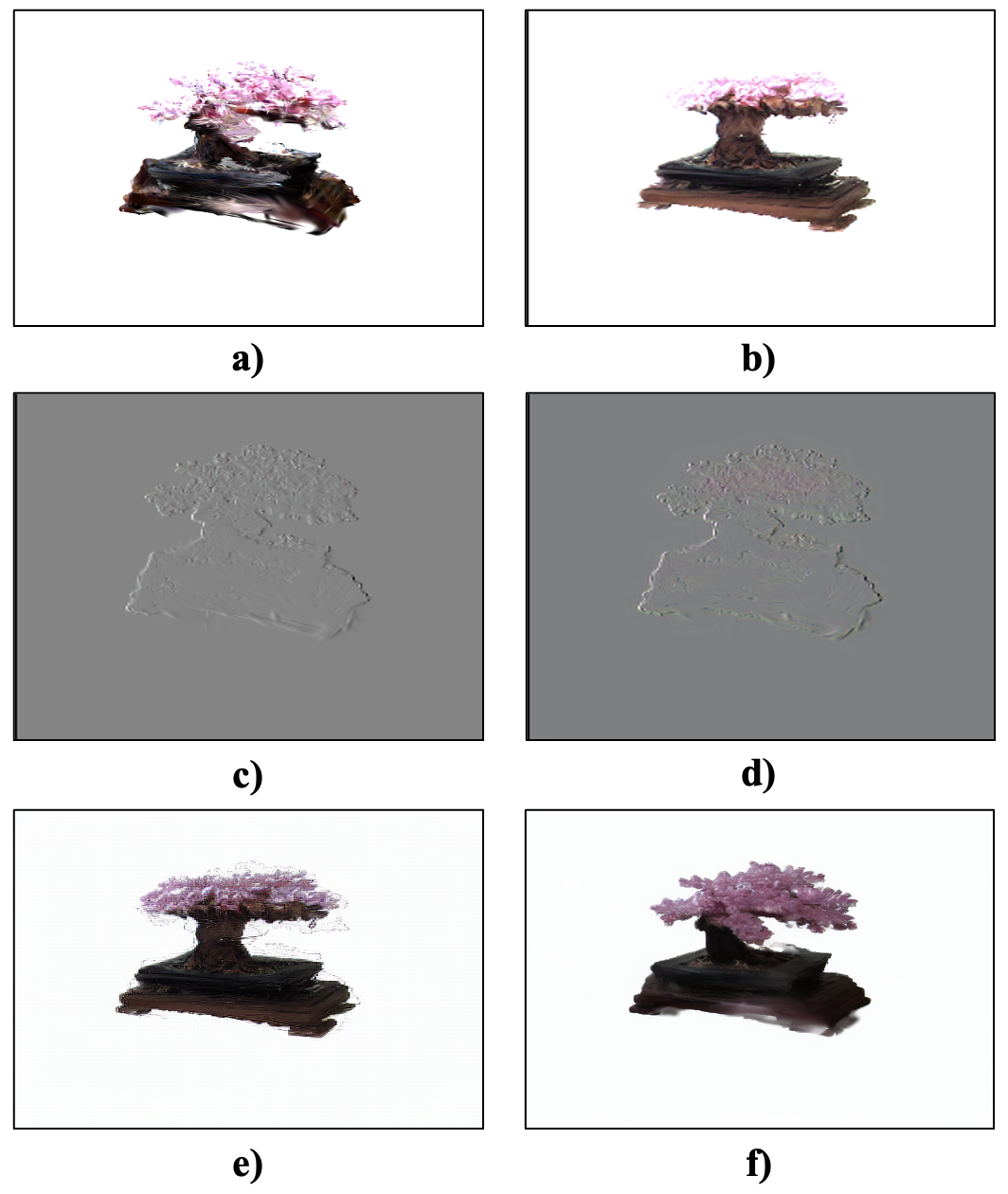}
    \caption{Pseudo view generation via LL-domain diffusion. Given corrupted LL and LH subbands (upsampled for better visualization) at a) and c), our framework provides the corresponding repairs at b) and d). Through the Inverse DWT, we generate a pseudo-sample at e). This bypasses RGB-domain diffusion at f), while providing comparable results.}
    \label{fig_fake}
\end{figure}

Finally, in the \textit{Fine Training} (d) stage, the coarse model $\mathcal{G}_{c}$ is refined into $\mathcal{G}_{f}$. During this process, $\mathcal{D}$, which is now frozen, repairs the LL renders of $\mathcal{G}_{c}$ from novel viewpoints, which are especially corrupted due to the sparse reference views. Similarly, the frozen $\mathcal{U}$ repairs the HF subbands. The repaired outputs of both are mapped back to the RGB domain through the Inverse DWT. Figure \ref{fig_fake} illustrates the use of the Inverse DWT to map the repairs back to original resolution.

Alongside actual references, the resulting IDWT reconstructions serve as pseudo references in the fine optimization step, thus emulating artifact-free dense-view supervision.

\subsection{Random Masking for Efficient Dataset Creation} \label{sec_orm}

To simulate corrupted patterns for \textit{Dataset Creation}, many state-of-the-art methods \cite{GaussianObject, RI3D, 4DSloMo} adopt a leave-one-out (LOO) strategy. This involves training $N$ separate 3DGS models ${\mathcal{G}_{d1},...,\mathcal{G}_{dN}}$, each constructed using all but one of the $N$ sparse reference views. The excluded view serves as the reference, while the render from same viewpoint is the corrupted counterpart. While effective at simulating corrupted patterns, training $N$ separate 3DGS models solely for this purpose is highly inefficient. Therefore, we introduce the online random masking (ORM) strategy. As shown in Figure \ref{fig_arch}, it only requires training a single $\mathcal{G}_d$, which is optimized using the typical pixel-wise rendering loss $\mathcal{L}_{\text{3DGS}}$ \cite{3DGS}. However, the references at index $n \in [1,N]$, $\mathbf{X}^{\text{gt}}_{n}$, are randomly masked with a binary mask $\mathbf{M}$. It consists of $n_{\mathbf{m}}$ 0-valued regions, each denoted as $\mathbf{m}$, to only mask certain regions of $\mathbf{X}^{\text{gt}}_{n}$. Each region $\mathbf{m}$ drifts according to sinusoidal displacements during training to generate diverse corruption patterns for $\mathcal{D}$. $\mathbf{M}$ is applied differently to each $\mathbf{X}^{\text{gt}}_{n}$ in the dataset, and simulates lack of coverage while using all $N$ views at a time, thus bypassing the LOO strategy. 

\section{Experiments}
\label{sec:exp}

\subsection{Datasets \& Implementation Details}

\textbf{Datasets \& Metrics}. The object reconstruction performance of WaveletGaussian is evaluated by measuring the quality of novel view renderings on held-out views of the Mip-NeRF 360 \cite{Mipnerf360} and OmniObject3D \cite{OmniObject3D} datasets, using the PSNR, SSIM and LPIPS metrics. Additionally, the end-to-end training time is recorded.

\textbf{Implementation Details}. Our implementation is built upon GaussianObject \cite{GaussianObject}. Different to ours, it leverages the LOO strategy and RGB-domain $\mathcal{D}$ for novel view repairs. Firstly, to replace LOO, we adopt the ORM strategy described at Section \ref{sec_orm}. During training $\mathcal{G}_d$, we use a mask $\mathbf{M}$ with $n_\mathbf{m} = 10$ masking regions, the total area of which covers 50\% of the object. Secondly, similar to GaussianObject, we leverage a pre-trained ControlNet \cite{ControlNet} for $\mathcal{
D}$. All training parameters remain the same, except $\mathcal{D}$ is fine-tuned on LF corrupted-clean pairs. The HF-repairing $\mathcal{U}$ processes concatenated HF subbands and is terminated based on early stopping to prevent overfitting.

\subsection{Quantitative Results} \label{sec_quant}

We present the quantitative results in Table \ref{tab_quant}. Generally, compared to the closest baseline, GaussianObject \cite{GaussianObject}, our proposed method achieves a 0.3-0.5 dB increase in PSNR and cuts the overall training time roughly by 40\%.

\subsection{Ablation Studies} \label{sec_abs}

Table \ref{tab_abs} presents ablation results. Firstly, we replace the LOO strategy, utilized by the baseline, with two variations of the random masking strategy. Different from the Online RM strategy presented in Section \ref{sec_orm}, the Offline RM strategy does not incorporate drifting masks. The former achieves better PSNR because the more diverse corruption patterns make $\mathcal{D}$ more robust. Both strategy outperform LOO in training time due to training a single $\mathcal{G}_d$, and without performance reductions. Having incorporated ORM, we then use wavelet diffusion (``wavelet-$\mathcal{D}$'') for novel view repairs. This further decreases training time, but the PSNR suffers because $\mathcal{D}$ only rectifies the coarse LL subbands. Incorporating $\mathcal{U}$ to rectify HF subbands leads to the best results, at the cost of some minor additional training time.

\section{Conclusion}

We introduce WaveletGaussian, a sparse-view 3D Gaussian object reconstruction framework that leverages a wavelet-domain diffusion model for novel view repairs. The switch from RGB to lower-resolution wavelet domain significantly reduces overall training time, while enabling frequency-separated repairs without performance degradation, as supported by experimental results.




\textbf{Compliance with Ethical Standards}. This is a numerical simulation study for which no ethical approval was required.

\textbf{Acknowledgements}. The first author was supported by the Vingroup Science and Technology Scholarship Program for Overseas Study for Master’s and Doctoral Degrees.


\bibliographystyle{IEEEbib}
\bibliography{strings,refs}

\end{document}